\documentclass[runningheads]{llncs}

\usepackage{accv}

\usepackage{accvabbrv}

\usepackage{graphicx}
\usepackage{booktabs}

\usepackage{graphics}
\usepackage{caption}
\usepackage{mathptmx}

\usepackage{amsmath}
\usepackage{amssymb}
\usepackage[protrusion=true,expansion=true]{microtype}
\usepackage{cite}
\usepackage{algorithm}
\usepackage{algpseudocode}
\usepackage{float}
\usepackage{comment}

\usepackage[accsupp]{axessibility}  

\usepackage{hyperref}

\usepackage{orcidlink}

\begin{document}

\title{LHM-Humanoid: Long-Horizon Human Motion Control for Continuous Object Transport in Cluttered Scenes} 

\titlerunning{LHM-Humanoid}

\author{Haozhuo Zhang\inst{1} \and
Jingkai Sun\inst{2} \and
Michele Caprio\inst{3} \and
Angelo Cangelosi\inst{1} \and
Jian Tang\inst{4} \and
Shanghang Zhang\inst{5} \and
Qiang Zhang\inst{6}\textsuperscript{*} \and
Wei Pan\inst{7}\textsuperscript{*}}

\authorrunning{H.~Zhang et al.}

\institute{The University of Manchester, Manchester, UK,
\email{haozhuo.zhang@postgrad.manchester.ac.uk},
\email{angelo.cangelosi@manchester.ac.uk} \and
The University of Hong Kong, Hong Kong, China,
\email{jingkaisun@connect.hku.hk} \and
University of Warwick, Coventry, UK,
\email{Michele.Caprio@warwick.ac.uk} \and
X-Humanoid, \email{jian.tang@x-humanoid.com} \and
Peking University, Beijing, China,
\email{shanghang@pku.edu.cn} \and
University of Science and Technology of China, Hefei, China,
\email{jonyzhang@ustc.edu.cn} \and
Newcastle University, Newcastle upon Tyne, UK,
\email{wei.pan2@newcastle.ac.uk}}

\maketitle
\begingroup
\renewcommand{\thefootnote}{*}
\footnotetext{Corresponding authors.}
\endgroup

\begin{abstract}
  Physics-based human motion control can make a simulated character walk, sit, and manipulate objects with high physical realism. Almost always, though, this happens in short, isolated clips that are re-initialized between interactions. We instead aim for continuous, reset-free long-horizon motion: a physically simulated humanoid that repeatedly walks to a displaced object, lifts it with a balanced whole-body posture, carries it past obstacles, and places it at a goal, over and over within a single uninterrupted take. The hard part is not any individual motion but the transitions between them. Without a reset, each cycle must end in a state that both leaves the object just placed undisturbed and lets the next cycle begin, yet every placement leaves the character off-balance in a non-canonical pose where naive end-to-end reinforcement learning fails. Our key idea is to treat this handoff as a two-sided problem of recoverability: the character must disengage from the object it just placed so the prior success is preserved, and settle into a state from which a balanced continuation exists. Instead of engineering a transition by hand, we learn to shape where each cycle ends so that it lands in this recoverable region. We introduce LHM-Humanoid. One goal-conditioned controller completes a fetch--carry--place cycle and, through a learned release-and-retreat behavior, steers its terminal state into this region; a second controller then takes over from the resulting state distribution. Both are regularized by an adversarial motion prior and distilled into a single goal-conditioned policy that runs the whole sequence as one reset-free rollout. Across 350 cluttered layouts spanning four room types, LHM-Humanoid produces far more successful and stable long-horizon motion than end-to-end RL, hierarchical RL, and prior physics-based human-scene-interaction methods, on both seen and unseen scenes. Trained only on two-object episodes, it extrapolates zero-shot to five-object sequences, where the baselines collapse to near-zero success. The resulting behavior can also be driven by egocentric vision and language. Project page: \url{https://haozhuo-zhang.github.io/LHM-Humanoid}.
\end{abstract}

\section{Introduction}

\begin{figure*}[!t]
  \centering
  \includegraphics[width=0.719\linewidth]{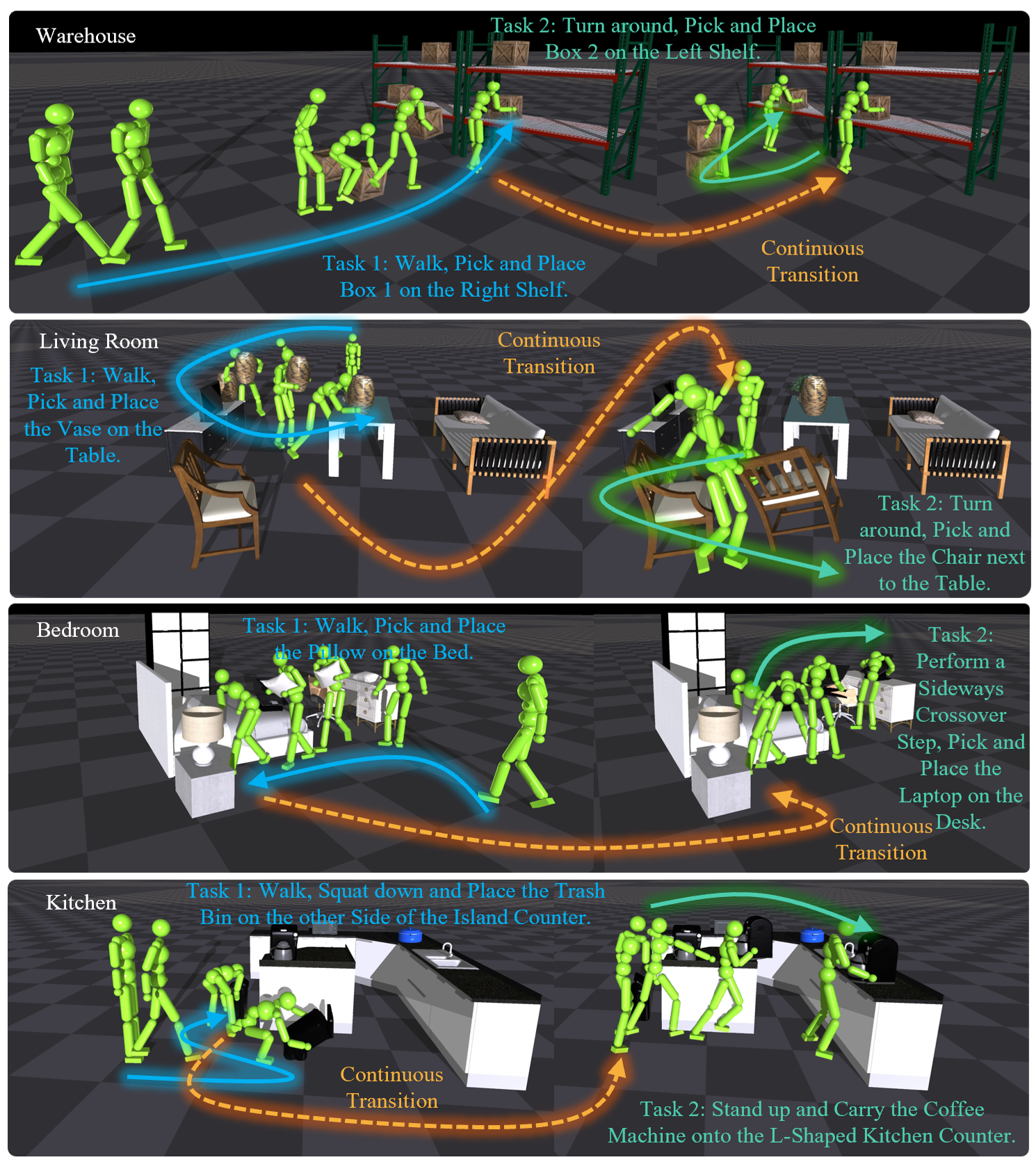}
  \caption{\textbf{Overview of LHM.}
    Our approach produces long-horizon whole-body human motion across four distinct environments: Warehouse, Living Room, Bedroom, and Kitchen. A single policy drives a physically simulated character to continuously walk to, pick, carry, and place multiple objects in unseen cluttered scenes, adapting to varied layouts and balance constraints without intermediate resets.}
  \label{fig:first-img}
\end{figure*}

Generating physically plausible human motion in which a character navigates a scene, interacts with objects, and rearranges its surroundings is a long-standing goal of character animation and embodied simulation. Prior work has made substantial progress in physics-based motion control~\cite{peng2021amp, tessler2023calm, luo2023perpetual, zhu2023neural}, human-scene interaction~\cite{starke2019neural, hassan2023synthesizing, xiao2023unified, pan2024synthesizing}, and object manipulation~\cite{xu2024humanvla, wang2023physhoi, xie2023hierarchical}. However, most existing settings simplify either the task horizon (to single-step or single-object interactions) or the scene distribution (training and evaluating within a single or fixed scene set)~\cite{wang2024skillmimic, xu2025intermimic, pan2025tokenhsi}.

We push physics-based motion control into a much more demanding regime: continuous, reset-free long-horizon whole-body motion, in which a simulated character transports multiple objects across a cluttered scene. Consider a room where several objects sit in the wrong place, such as a laptop left on a bed or a trash bin blocking a walkway. Producing believable motion here calls for sustained coordination of locomotion, whole-body manipulation, and object transport over a long horizon. The character must repeatedly handle misplaced objects under balance and obstacle constraints, and recover on its own from the awkward body configurations left after each placement, all in one uninterrupted physically simulated take. Three requirements together separate our setting from prior motion-control work: long-horizon whole-body interaction without resets, cross-scene generalization, and production of the whole sequence from a single unified controller instead of stitched per-skill clips.

Our central observation reframes what makes this hard. The difficulty is not producing any single motion, such as walking, grasping, or placing, but composing them across the seams. In a reset-free episode, the state where one cycle ends becomes the state where the next begins, so stability depends on two things at once: each cycle must end without disturbing the object it just placed, and it must leave the character somewhere a continuation is still feasible. The problem is that every placement leaves the character in a non-canonical, off-balance pose, and naive end-to-end RL breaks down at these transitions. Our answer is a learning principle rather than a scripted maneuver: make the termination distribution recoverable. We train one controller to steer the end of each cycle into the recoverable region, the states from which a balanced continuation provably exists, and a second controller to start from that induced distribution, so the cycles compose without a reset. The hard step is not the release-and-retreat behavior itself, but recognizing recoverability as the governing principle and learning to satisfy it under contact-rich dynamics across hundreds of scenes.

To support research on this problem, we build a dataset of 350 distinct scenes and tasks across four room types, with wide variation in object identities, clutter arrangements, and start/goal configurations. Importantly, it is a task-and-scene benchmark rather than a demonstration dataset: it specifies layouts, object sets, and goals, but provides no scene-specific ground-truth motion. This rules out strict motion mimicry and forces the agent to acquire behavior from task objectives. The large variation within and across room categories also makes room-specific controllers and fixed skill libraries inadequate, so genuine generalization is required.

We introduce LHM-Humanoid, a control pipeline built around a single unified controller. Two goal-conditioned controllers are trained with physics-based RL and an adversarial motion prior~\cite{peng2021amp} for human-likeness. Teacher~1 produces the first fetch-carry-place motion and learns a termination-shaping behavior, realized as release-and-retreat, that drives its terminal state into the recoverable region. Teacher~2 then starts from that induced non-canonical distribution and completes the next cycle. Both are distilled into one unified goal-conditioned controller with DAgger~\cite{ross2011reduction}, which produces seamless long-horizon motion, including the inter-cycle handoff, as one continuous, reset-free rollout. This controller can be further distilled into a model~\cite{xu2024humanvla, ding2025humanoid} conditioned on egocentric RGB and natural language.

\noindent\textbf{Contributions:}
\begin{itemize}
  \item \textbf{A new motion-control problem and perspective}: continuous, reset-free long-horizon whole-body multi-object transport. We identify that the bottleneck is the inter-cycle handoff, governed by whether each cycle ends in a recoverable state, rather than the individual motions. In this regime end-to-end RL fails to produce stable motion and prior physics-based methods break down.
  \item \textbf{Learned viability-aware termination shaping}: instead of hand-engineering inter-cycle transitions, we learn a termination behavior (release-and-retreat) that drives each cycle's terminal distribution into the recoverable region, so the character leaves the object it just placed undisturbed and lands in the set of states from which the next cycle can start. We pair it with a recovery controller trained on exactly that induced distribution, regularize it with an adversarial motion prior, and distill it with DAgger into a single controller.
  \item \textbf{Strong results and a controllable extension}: LHM-Humanoid produces more successful and stable long-horizon interaction than end-to-end RL, hierarchical RL, and prior physics-based human-scene-interaction baselines on seen and unseen scenes. It extrapolates zero-shot from two-object to five-object sequences and can be driven by egocentric RGB and language. We support these results with a diverse testbed of 350 long-horizon cluttered scenes across four room types.
\end{itemize}

\section{Related Work}

\subsection{Human Motion Modeling and Physically-Based Human-Scene Interaction}
Motion synthesis spans kinematic approaches using VAEs, transformers, and diffusion models~\cite{li2023object, huang2023diffusion, jiang2024scaling, zhao2023synthesizing, jiang2023motiongpt, lin2023motion, cai2021unified} and physics-based methods that learn controllers under simulation constraints~\cite{xiao2023unified, luo2023perpetual, tessler2023calm, zhu2023neural, sferrazza2024humanoidbench, xie2023hierarchical, pan2024synthesizing, wang2023physhoi, braun2024physically}. Physics-based controllers have been advanced through DeepMimic-style imitation~\cite{peng2018deepmimic}, adversarial motion priors~\cite{peng2021amp}, discrete latent skills~\cite{zhu2023neural}, reusable skill embeddings~\cite{peng2022ase}, language conditioning~\cite{juravsky2022padl}, and expressive control~\cite{tessler2023calm}.

Human-scene interaction (HSI) further addresses contact-rich behaviors including sitting, lying, and object use, with representative works such as InterPhys~\cite{hassan2023synthesizing}, InterScene~\cite{pan2024synthesizing}, and UniHSI~\cite{xiao2023unified}. Earlier physics-based HSI approaches relied on IK, optimal control, or handcrafted controllers~\cite{elkoura2003handrix, kim2000neural, jain2011controlling, zhao2013robust}, while recent methods leverage deep RL~\cite{cui2024anyskill, dou2023c, hassan2023synthesizing, serifi2024vmp, yao2024moconvq, zhu2023neural} to acquire richer skills such as dribbling~\cite{wang2024skillmimic}, skateboarding~\cite{liu2017learning}, and tool use~\cite{yang2022learning}. Recent scaling efforts including SkillMimic, InterMimic, and TokenHSI~\cite{wang2024skillmimic, xu2025intermimic, pan2025tokenhsi} further expand physics-based interaction learning, but most controllers remain specialized to narrow task distributions and short, isolated interaction clips. Our work builds on these foundations but synthesizes long-horizon whole-body human-scene interaction as one continuous, reset-free sequence across diverse cluttered scenes.

\subsection{Room-Scale Object Rearrangement in Cluttered Scenes}
Room-scale rearrangement has been extensively studied in embodied AI and robotics~\cite{yenamandra2023homerobot, deitke2022️}, with benchmarks such as Visual Room Rearrangement~\cite{weihs2021visual} and OVMM~\cite{yenamandra2023homerobot}, and LLM-based planning approaches~\cite{wu2023tidybot}. However, these settings typically evaluate kinematic or simplified-embodiment agents and therefore do not model the physics-based whole-body challenges central to our problem---dynamic balance, fall avoidance, whole-body coordination, and contact-rich interaction---so they are not directly comparable to physically simulated motion control. HumanVLA~\cite{xu2024humanvla} advances toward heavier object manipulation with a physically simulated human character, but episodes remain relatively short. Our work instead produces long-horizon whole-body object transport as one continuous, reset-free sequence across 350 diverse cluttered scenes; as an extension, we further condition the resulting motion on language and egocentric vision, relating to language-conditioned motion synthesis~\cite{chen2023executing, jiang2023motiongpt, zhao2022compositional} and vision-language-action policies~\cite{zitkovich2023rt, xu2024humanvla, ding2025humanoid, wang2025unified, driess2025knowledge}.

\begin{figure*}[!t]
  \centering
  \includegraphics[width=1.0\linewidth]{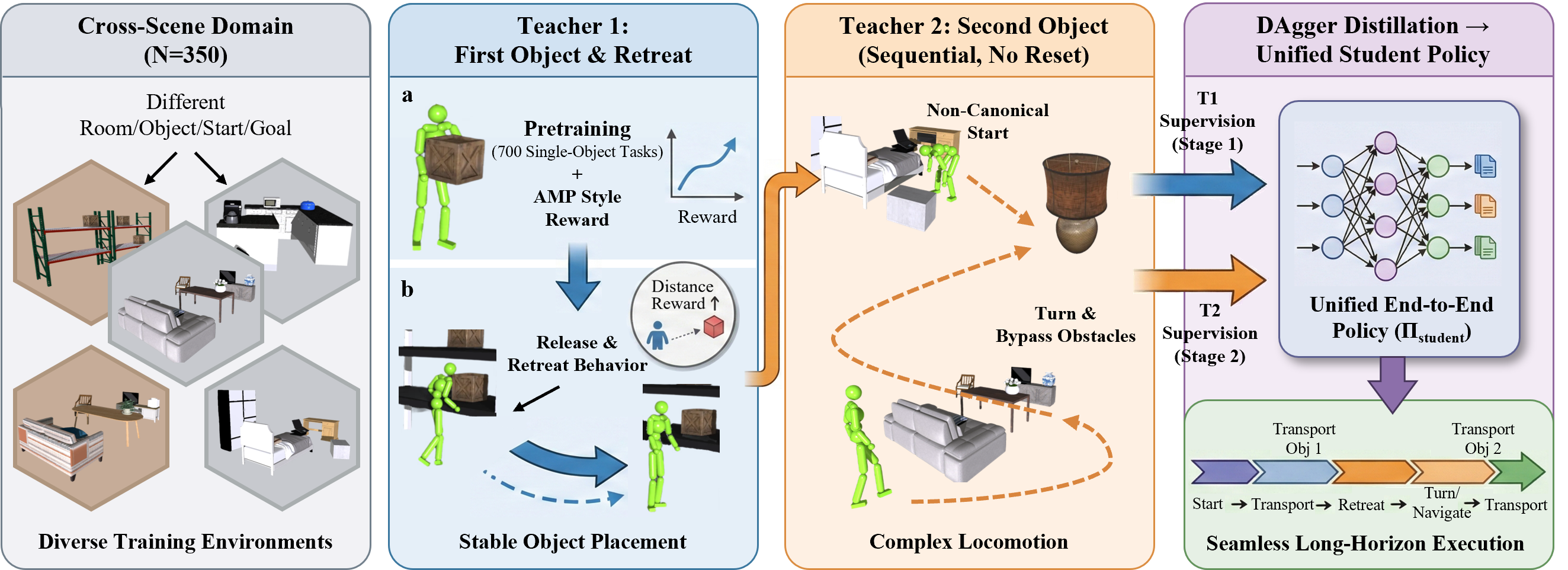}
  \caption{\textbf{Overview of the LHM learning framework.}
    The pipeline consists of three stages. \textbf{Left}: A diverse dataset of 350 scenes provides training coverage across rooms, objects, and configurations.
    \textbf{Middle}: Two goal-conditioned teacher controllers cover a continuous two-object episode without reset. Teacher~1 completes the first walk--pick--carry--place cycle and executes a release-and-retreat transition to a stable state; Teacher~2 starts from that non-canonical terminal state and completes the next object cycle.
    \textbf{Right}: Both teachers are distilled into one unified goal-conditioned student policy via DAgger, yielding a single network that executes each cycle and the inter-cycle transitions as one continuous, reset-free rollout.}
  \label{fig:main_img}
\end{figure*}

\section{Method}
\label{sec:method}

We formalize reset-free long-horizon whole-body object transport in cluttered scenes and describe our three-stage training pipeline (Figure~\ref{fig:main_img}). It is natural to view each fetch-carry-place cycle as a temporally extended action, or option~\cite{sutton1999between},
\begin{equation}
  o=\langle\mathcal{I}_o,\;\pi_o,\;\beta_o\rangle,
  \label{eq:option}
\end{equation}
with an initiation set $\mathcal{I}_o$, an intra-option policy $\pi_o$, and a termination condition $\beta_o$. We use this only as an analytical lens: control and reward stay at the primitive timestep, so a full episode can be described as a trajectory over the semi-MDP these cycles induce. What makes the reset-free setting distinctive is that the state where one option ends must fall inside the initiation set of the next. A central part of our design is therefore to shape $\beta_o$ through release-and-retreat, rather than treat it as a fixed boundary. Because cross-scene generalization is central, all policies are trained jointly over 350 diverse scenes and tasks. There is no scene-specific ground-truth motion, and the configurations vary widely, so fixed skill composition is fragile and direct end-to-end RL on this long-horizon, contact-rich task tends not to converge. We therefore use a dual-teacher mechanism: Teacher~1 shapes its terminal state into the recoverable region, and Teacher~2 starts from that induced distribution to complete the next cycle. We then apply DAgger~\cite{ross2011reduction} to distill both into a single end-to-end student, with a further distillation step producing a vision-language-action (VLA) model for RGB-and-language conditioned control.

\subsection{Teacher Policy 1: First-Object Transport with Release-and-Retreat}
\label{subsec:teacher-policy-1}

Teacher~1 completes the first fetch-carry-place cycle. Crucially, it must end inside the recoverable region: it has to leave the object it just placed undisturbed while putting its terminal distribution inside Teacher~2's initiation set. We therefore treat termination shaping, the requirement that each cycle ends recoverable, as a first-class training objective rather than an afterthought, and realize it through a learned release-and-retreat behavior. If Teacher~1 instead ends off-balance or entangled, Teacher~2 inherits an out-of-distribution state and the error compounds over the rest of the episode. Deciding where Teacher~1 stops is thus what makes the later cycles learnable at all. It also shifts the burden from designing the transition by hand to specifying one property, recoverability of the terminal state, that the policy can then optimize end-to-end.

\paragraph{Pretraining on single-object tasks.}
We first pretrain Teacher~1 to acquire robust single-object fetch-carry-place behaviors. Each two-object episode is split into two single-object goals, giving 700 single-object tasks that expose the policy to diverse start states, clutter configurations, and spatial layouts. To encourage human-like motion, we add an Adversarial Motion Prior (AMP)~\cite{peng2021amp} style reward $r_{\text{style}}$, computed from a learned discriminator that separates human-like from non-human-like state transitions. We optimize the total reward $r_{\text{total}} = r_{\text{task}} + r_{\text{style}}$, where $r_{\text{task}}$ is a shaped task reward for locomotion-to-grasp, grasping, carrying, and placement. The full reward definitions are in the supplementary material.

\paragraph{Release-and-retreat fine-tuning.}
After pretraining, we fine-tune Teacher~1 to drive its terminal state into the recoverable region. The configurations we target form a recoverable (viable) region~\cite{aubin1991viability}, the states from which a balanced continuation exists. Concretely, a recoverable terminal state---which also serves as the initiation set from which the second cycle begins---is one where the character has fully released and moved clear of the object it just placed, so that starting the next cycle cannot disturb it, while staying balanced and well-positioned to set off toward the next target. This is closely related to the capturable states in bipedal push recovery~\cite{pratt2006capture}, and fine-tuning aims to make Teacher~1 end inside this set, within Teacher~2's domain of attraction. Membership in the viable set is intractable to optimize directly under contact-rich dynamics, so we use release-and-retreat as a tractable surrogate: a reward that grows with the (capped) torso- and hand-to-object distances after placement, and that stays masked until the object is within a small radius of its goal. Rewarding separation only after placement steers the character into this region and holds it there until handoff, rather than merely stepping back. Because the term activates only after a successful placement, it never competes with the task reward during transport, so it does not distort the carrying trajectory; it acts purely as a terminal-state regularizer that reshapes the distribution Teacher~2 must handle. Algorithm~\ref{alg:teacher_1_reward} gives the exact form. The reward grows with the capped torso- and hand-to-object distances after placement (lines~1--9), and stays masked until the object is within a small radius of its goal (lines~10--13), which prevents a premature retreat. The final reward is a weighted combination of the two distance terms (line~14).

\begin{algorithm}[t]
  \caption{Release-and-Retreat Reward Function}
  \label{alg:teacher_1_reward}
  \footnotesize
  \begin{algorithmic}[1]
    \State $d_{\text{hand2object}} \gets \min\!\left(d_{\text{hand2object}}^{\text{left}},\, d_{\text{hand2object}}^{\text{right}}\right)$
    \State $r_{\text{root2object}} \gets 1 - \exp(-\alpha \cdot d_{\text{root2object}})$
    \State $r_{\text{hand2object}} \gets 1 - \exp(-\alpha \cdot d_{\text{hand2object}})$
    \If{$d_{\text{root2object}} > \delta_r$} \State $r_{\text{root2object}} \gets 1$ \EndIf
    \If{$d_{\text{hand2object}} > \delta_h$} \State $r_{\text{hand2object}} \gets 1$ \EndIf
    \If{$d_{\text{object2goal}} \geq \tau_{\text{object2goal}}$}
      \State $r_{\text{root2object}} \gets 0$
      \State $r_{\text{hand2object}} \gets 0$
    \EndIf
    \State $\text{reward} \gets \beta \cdot r_{\text{root2object}} + (1-\beta) \cdot r_{\text{hand2object}}$
  \end{algorithmic}
\end{algorithm}

\subsection{Teacher Policy 2: Recovery Locomotion and Next-Object Transport from Non-Canonical States}

Teacher~2 takes over immediately after Teacher~1 completes its release-and-retreat, continuing the episode without any environment reset to complete the next-object fetch-carry-place cycle.

The primary challenge is that the handoff state is non-canonical. After the first placement and retreat, the character may exhibit diverse postures---crouched, leaning, with unevenly spaced feet or displaced hands---and can be oriented arbitrarily, possibly facing away from the next target. The scene may also be partially obstructed by the previously placed object, making subsequent navigation sensitive to incidental contacts.

Teacher~2 therefore solves a compound problem: recovery locomotion to regain a stable walking posture, reorientation toward the next target, obstacle-aware navigation through cluttered passages, and whole-body interaction to grasp, carry, and place the next object. To train it under this broadened distribution, each episode first runs Teacher~1 to completion, then rolls out Teacher~2 from the resulting state and optimizes it with the same goal-conditioned single-object objective used in pretraining. The reward is reused, but the effective training distribution is much harder, because the no-reset setting induces it. This encourages robustness to off-nominal initial states and implicit obstacle-aware navigation without an explicit planner. The design encodes a recovery-from-non-canonical-state mechanism rather than per-object specialization, so it is not tied to the number of objects, as confirmed by zero-shot extrapolation to longer sequences (Sec.~\ref{subsec:quant-results}).

\subsection{Distillation into a Unified End-to-End Policy}
\label{subsec:distillation}

The two teachers are distilled into a single goal-conditioned student that produces every cycle and the transitions between them as one continuous, reset-free rollout. A unified controller is essential here, because it must handle the entire state distribution induced by sequential, contact-rich interaction across diverse scenes. Direct end-to-end RL fails to converge in this setting, due to intractable credit assignment and exploration over long horizons, whereas DAgger-based distillation converges reliably and generalizes across scenes and to longer multi-object settings. Imitation-learning theory predicts this. Naive behavior cloning suffers compounding covariate shift, with a performance gap that grows quadratically in the horizon, $\mathcal{O}(T^2\epsilon)$, whereas aggregating data under the student's own induced state distribution gives a no-regret bound that is linear in the horizon,
\begin{equation}
  J(\pi) \le J(\pi^{*}) + u\,T\,\epsilon_N + \mathcal{O}(T\,\delta_N),
  \label{eq:dagger}
\end{equation}
where $\epsilon_N$ is the average online imitation loss, $u$ upper-bounds the cost-to-go difference, and the regret term $\delta_N\!\to\!0$ as data aggregation proceeds~\cite{ross2011reduction}. The linear dependence on $T$ is exactly what a reset-free, long-horizon setting needs, and it matches the gentle degradation of our policy as the number of objects grows (Sec.~\ref{subsec:quant-results}).

We use DAgger~\cite{ross2011reduction} for distillation. During data aggregation, Teacher~1 supervises the student through the first-object cycle (locomotion-to-grasp, carry, place, release, retreat), and Teacher~2 supervises through the next-object cycle. At each timestep, a finite-state machine (FSM) selects the supervising teacher according to the stage of the episode, which gives stage-appropriate supervision while still training a single unified policy.

\subsection{Extension: Distillation into an End-to-End VLA Model}
\label{subsec:extension-vla}

As an extension, we distill the unified student into a vision-language-action (VLA) model conditioned on natural language instructions and egocentric RGB observations. The VLA model replaces privileged oracle state inputs with two perception-grounded modalities: a text instruction specifying the current placement goal, and the character's first-person RGB stream. Following the same DAgger principle, the unified student provides action supervision, and the VLA model learns to imitate these actions from RGB and language alone. At test time, one can issue consecutive natural language instructions, and the VLA policy produces each fetch-carry-place cycle, with release-and-retreat transitions, within the same episode. This enables long-horizon instruction following with a single end-to-end model. Here the language token is not a redundant goal specifier: under first-person occlusion it is the only cue that tells apart visually similar movable objects (\eg, a pot vs.\ a coffee-maker on the same stand), and removing it collapses long-horizon success (Sec.~\ref{subsec:robustness}).

\section{Experiments}

\subsection{Dataset and Evaluation Metrics}
\label{subsec:dataset}

Our dataset covers four room types---bedroom, living room, kitchen, and warehouse---comprising 350 diverse cluttered scene layouts and long-horizon object-transport episodes. Scenes contain 79 distinct objects, of which 25 are movable placement targets. Each episode requires completing two consecutive fetch-carry-place cycles in a single continuous rollout (no resets), accompanied by two natural language instructions describing the respective placement goals. To evaluate generalization, we curate 66 unseen tasks with distinct layouts, objects, and instructions as an out-of-distribution benchmark.

\paragraph{Benchmark statistics and diversity.}
Variation in the benchmark is multi-axis and goes well beyond textures or backgrounds. There are 79 objects (25 movable targets) with large shape dispersion, roughly 4--10 obstacles per scene at varying footprints, heights, and placements, and per-room object initial/goal radii up to about $1.5$\,m with held-out scene-centroid shifts up to about $0.5$\,m, on the order of $2$--$3$ character steps. The obstacle layouts are what motivate obstacle-aware carrying and release-and-retreat, rather than straight-line transport. Because the held-out (initial, goal) configurations and centroid shifts are well beyond a single step, success on this split reflects genuine spatial generalization rather than memorization. For human-likeness, AMP uses reference motions from OMOMO~\cite{li2023object} and SAMP~\cite{hassan2021stochastic}.

Performance is reported via \textbf{Success~1} and \textbf{Success~2} (per-cycle success rates), \textbf{Success~All} (both cycles completed in one episode), and \textbf{Dist~1}/\textbf{Dist~2} (final placement error in meters). A cycle is counted as successful when the object root position is within 0.2\,m of the goal position. Higher success rates and lower distances indicate better performance. Unless otherwise noted, results are averaged over three random seeds, with a typical variance of about $\pm 3.2$ percentage points on Success~All.

\subsection{Quantitative Results}
\label{subsec:quant-results}

We report results on the 350 training tasks, 66 unseen tasks, and the VLA extension. \textbf{RR} denotes the release-and-retreat objective in Teacher~1; \textbf{T} denotes the dual-teacher design; \textbf{S} denotes the distilled student.

\paragraph{Baselines.}
We compare against End-to-End RL, Curriculum RL (the degenerate single-teacher case), Hierarchical RL (five low-level skills---walk, reach-and-grasp, lift, obstacle-aware carry, place-and-release---switched by a privileged FSM), and the physics-based HSI methods HumanVLA~\cite{xu2024humanvla}, InterMimic~\cite{xu2025intermimic}, and TokenHSI~\cite{pan2025tokenhsi}, alongside our RR/AMP ablations. All methods in Tables~\ref{tab:main_results}--\ref{tab:three-object} share identical observation and action interfaces and were re-run with matched environments and compute, so performance differences reflect learning strategy rather than input/output disparities. Exact per-baseline configurations are provided in the supplementary material.

\paragraph{Main results on 350 tasks.}

\begin{table*}[htbp]
  \centering
  \caption{Results on 350 training tasks.}
  \label{tab:main_results}
  \begin{tabular}{lccccc}
    \toprule
    Method & Success 1 $\uparrow$ & Success 2 $\uparrow$ & Success All $\uparrow$ & Dist 1 (m) $\downarrow$ & Dist 2 (m) $\downarrow$ \\
    \midrule
    End-to-End RL          & 2.3\%  & 0.0\%  & 0.0\%  & 2.05 & 1.97 \\
    Curriculum RL          & 88.3\% & 48.5\% & 47.2\% & 0.26 & 1.01 \\
    Hierarchical RL        & 37.5\% & 25.0\% & 20.8\% & 0.78 & 1.13 \\
    HumanVLA~\cite{xu2024humanvla}  & 42.3\% & 36.7\% & 29.9\% & 0.65 & 0.97 \\
    InterMimic~\cite{xu2025intermimic} & 20.6\% & 9.8\%  & 7.5\%  & 1.32 & 1.82 \\
    TokenHSI~\cite{pan2025tokenhsi}   & 40.5\% & 31.5\% & 27.6\% & 0.58 & 1.07 \\
    LHM w/o RR    & 81.1\% & 60.5\% & 56.2\% & 0.52 & 0.59 \\
    LHM w/o AMP   & 32.9\% & 14.8\% & 10.4\% & 1.09 & 1.64 \\
    \textbf{LHM-S}& 87.4\% & 72.6\% & 71.1\% & 0.28 & 0.50 \\
    \textbf{LHM-T}& \textbf{88.8\%} & \textbf{72.9\%} & \textbf{72.4\%} & \textbf{0.25} & \textbf{0.48} \\
    \bottomrule
  \end{tabular}
\end{table*}

Table~\ref{tab:main_results} reveals systematic failure modes across baseline paradigms. End-to-End RL collapses to 0\% Success~All: credit assignment and exploration are intractable when a policy must discover multi-stage, contact-rich behaviors from sparse rewards over long horizons. Curriculum RL shows that a simple curriculum is not enough, with strong early stages (Success~1: 88.3\%) but compounding degradation (Success~All: 47.2\%). Hierarchical RL and TokenHSI~\cite{pan2025tokenhsi} are brittle because their low-level controllers and primitives cover narrow action manifolds, so small changes in object pose or clutter create out-of-distribution conditions that compound across stages. InterMimic~\cite{xu2025intermimic} relies only on task- and scene-aligned supervision, so it transfers poorly to our cluttered benchmark. HumanVLA~\cite{xu2024humanvla}, though unified, degrades over long horizons (Success~All: 29.9\%) under sequential distribution shift.

Our method addresses these issues with three complementary mechanisms. Release-and-retreat enforces a stable terminal state after each placement, which reduces error propagation; dual-teacher training broadens state coverage beyond canonical trajectories; and DAgger distillation yields a single unified policy that generalizes consistently. LHM-T achieves the best Success~All (72.4\%) with the lowest placement errors (0.25/0.48\,m), and the distilled LHM-S (71.1\%) stays close to it. On these same teachers, naive behavior cloning reaches only $32.7\%$ Success~All ($70.6\%$/$48.3\%$), while DAgger reaches $71.1\%$ by aggregating labels on the student's own states. Ablations confirm that both RR (--16.2\%) and AMP (--62.0\%) are critical. Release-and-retreat pays off only after transport is learned: joint training from scratch reaches a Success~1 of only $10\%$, against $88.8\%$ after fine-tuning. Given a successful first placement, the next cycle then succeeds in $81.5\%$ of LHM-T episodes, compared with $53.5\%$ for Curriculum RL and $69.3\%$ without release-and-retreat, and the placed object is later displaced in $8.2\%$ of episodes rather than $26.0\%$.

\paragraph{Results on 66 unseen tasks.}

\begin{table}[htbp]
  \centering
  \caption{Results on 66 unseen tasks.}
  \label{tab:unseen_results}
  \begin{tabular}{lccccc}
    \toprule
    Method & Success 1 $\uparrow$ & Success 2 $\uparrow$ & Success All $\uparrow$ & Dist 1 (m) $\downarrow$ & Dist 2 (m) $\downarrow$ \\
    \midrule
    End-to-End RL          & 1.4\%  & 0.0\%  & 0.0\%  & 2.37 & 2.28 \\
    Curriculum RL          & 75.5\% & 40.4\% & 39.5\% & 0.36 & 1.11 \\
    Hierarchical RL        & 31.6\% & 21.8\% & 17.3\% & 0.97 & 1.39 \\
    HumanVLA~\cite{xu2024humanvla}  & 36.3\% & 31.3\% & 25.1\% & 0.72 & 1.16 \\
    InterMimic~\cite{xu2025intermimic} & 17.2\% & 8.2\%  & 5.9\%  & 1.47 & 2.03 \\
    TokenHSI~\cite{pan2025tokenhsi}   & 33.9\% & 26.1\% & 22.9\% & 0.67 & 1.21 \\
    LHM w/o RR    & 69.2\% & 52.4\% & 48.6\% & 0.62 & 0.60 \\
    LHM w/o AMP   & 27.2\% & 12.9\% & 9.0\%  & 1.27 & 1.86 \\
    \textbf{LHM-S}& 80.8\% & 64.9\% & 61.6\% & 0.37 & 0.54 \\
    \textbf{LHM-T}& \textbf{81.6\%} & \textbf{65.1\%} & \textbf{63.2\%} & \textbf{0.35} & \textbf{0.50} \\
    \bottomrule
  \end{tabular}
\end{table}

On unseen scenes (Table~\ref{tab:unseen_results}), all baselines degrade under distribution shift. Curriculum RL and Hierarchical RL overfit their staged or skill-level training distributions, so novel object poses or clutter push them outside their repertoire and errors compound across the horizon. HumanVLA, InterMimic, and TokenHSI also drop sharply, reflecting limited long-horizon state coverage and brittle motion-only or skill-token supervision. Our method generalizes more robustly. The release-and-retreat objective regularizes the inter-cycle handoff and limits error accumulation after each placement, and the dual-teacher setup exposes the student to a wider distribution of intermediate states than a single canonical trajectory. As a result, the unified policy stays stable under unseen layouts and contact patterns, re-aligns to goals after perturbations, and reaches 63.2\% Success~All at 0.35/0.50\,m placement error. The train$\rightarrow$held-out drop for LHM-T is modest, about $7.2/7.8/9.2$ points on Success~1/2/All (a relative drop of roughly $8$--$13\%$), and the policy still completes $63.2\%$ of full unseen episodes. Given the substantial spatial variation in Sec.~\ref{subsec:dataset} and the zero-shot longer-horizon results below, this points to genuine generalization rather than overfitting.

\paragraph{VLA extension results.}

\begin{table}[htbp]
  \centering
  \caption{VLA extension results (all methods distilled via the same DAgger pipeline).}
  \label{tab:vla_results}
  \begin{tabular}{lccccc}
    \toprule
    Method & Success 1 $\uparrow$ & Success 2 $\uparrow$ & Success All $\uparrow$ & Dist 1 (m) $\downarrow$ & Dist 2 (m) $\downarrow$ \\
    \midrule
    End-to-End RL          & 1.3\%  & 0.0\%  & 0.0\%  & 2.50 & 2.49 \\
    Curriculum RL          & 71.5\% & 41.1\% & 40.7\% & 0.37 & 1.18 \\
    Hierarchical RL        & 29.3\% & 20.7\% & 16.5\% & 1.04 & 1.45 \\
    HumanVLA~\cite{xu2024humanvla}  & 34.2\% & 29.8\% & 23.5\% & 0.70 & 1.26 \\
    InterMimic~\cite{xu2025intermimic} & 16.3\% & 8.4\%  & 6.2\%  & 1.64 & 2.22 \\
    TokenHSI~\cite{pan2025tokenhsi}   & 32.0\% & 27.0\% & 23.7\% & 0.72 & 1.31 \\
    LHM w/o RR    & 66.7\% & 52.2\% & 48.7\% & 0.60 & 0.68 \\
    LHM w/o AMP   & 25.3\% & 14.5\% & 10.6\% & 1.39 & 1.91 \\
    \textbf{LHM-S}& \textbf{76.5\%} & \textbf{65.4\%} & \textbf{63.7\%} & \textbf{0.42} & \textbf{0.56} \\
    \bottomrule
  \end{tabular}
\end{table}

Table~\ref{tab:vla_results} evaluates all methods under the same DAgger-based distillation into a VLA model, which isolates teacher policy quality as the only variable. Baseline VLA models degrade substantially, because their teachers either fail to acquire robust long-horizon behaviors or rely on narrow skill manifolds that do not survive the modality gap once oracle states are replaced by RGB-and-language inputs. Our VLA model instead keeps strong performance (Success~All: 63.7\%; Dist~1/2: 0.42/0.56\,m) and exceeds the strongest baseline by more than $20$ points in Success~All ($63.7\%$ vs.\ $40.7\%$). This shows that the sequential behaviors learned under our framework survive distillation and transfer to egocentric sensing with their spatial precision preserved. Ablations confirm that both RR and AMP are needed here too: removing RR drops Success~All to 48.7\% and removing AMP to 10.6\%.

\paragraph{Extension to more than two objects.}

\begin{table*}[htbp]
  \centering
  \caption{More-than-two-object extension results (models directly evaluated without additional fine-tuning).}
  \label{tab:three-object}
  \begin{tabular}{lcccccc}
    \toprule
    Method
    & Succ1 $\uparrow$
    & Succ2 $\uparrow$
    & Succ3 $\uparrow$
    & Succ4 $\uparrow$
    & Succ5 $\uparrow$
    & SuccAll $\uparrow$ \\
    \midrule
    End-to-End RL
    & 2.4\%  & 0.0\%  & 0.0\%  & 0.0\%  & 0.0\%  & 0.0\%  \\
    Curriculum RL
    & 90.4\% & 51.8\% & 34.3\% & 22.3\% & 6.7\%  & 1.7\%  \\
    Hierarchical RL
    & 38.6\% & 27.3\% & 16.6\% & 3.8\%  & 0.0\%  & 0.0\%  \\
    HumanVLA~\cite{xu2024humanvla}
    & 45.8\% & 40.5\% & 21.7\% & 5.6\%  & 1.2\%  & 0.1\%  \\
    InterMimic~\cite{xu2025intermimic}
    & 21.3\% & 10.2\% & 2.5\%  & 0.4\%  & 0.0\%  & 0.0\%  \\
    TokenHSI~\cite{pan2025tokenhsi}
    & 41.7\% & 32.7\% & 18.9\% & 4.9\%  & 1.1\%  & 0.1\%  \\
    LHM w/o RR
    & 82.1\% & 60.0\% & 21.6\% & 4.1\%  & 0.9\%  & 0.0\%  \\
    LHM w/o AMP
    & 33.6\% & 14.3\% & 2.0\%  & 0.9\%  & 0.0\%  & 0.0\%  \\
    \textbf{LHM-S}
    & \textbf{90.8\%} & \textbf{76.1\%} & \textbf{61.0\%} & \textbf{38.5\%} & \textbf{20.9\%} & \textbf{18.1\%} \\
    \bottomrule
  \end{tabular}
\end{table*}

Table~\ref{tab:three-object} evaluates models trained on two-object episodes and tested directly on longer sequences (up to five objects) without any fine-tuning, on a targeted benchmark of four representative scene types (warehouse, bedroom, living room, kitchen), each instantiated with multiple layouts and several sequential rearrangement tasks (construction details in the supplementary material).

All baselines degrade steeply with episode length and collapse to near-zero Success~All beyond three objects. Strong first-object numbers (\eg, Curriculum RL at 90.4\%) do not survive the long horizon and fall to single-digit or zero full-episode completion. Our method instead keeps a large advantage at every stage: LHM-S reaches 61.0\% on the third object, 38.5\% on the fourth, and 18.1\% overall on five-object sequences, which no baseline approaches. This gap shows that short-horizon competence does not translate into long-horizon robustness; error propagation and recovery failures dominate as sequences grow. A simple model makes the mechanism concrete; we intend it as an illustration of the trend rather than a derivation. If every cycle is entered from the recoverable region and completed with probability $q$, an $N$-object episode succeeds with probability
\begin{equation}
  \Pr\!\left[\text{Succ}_{\text{All}}^{(N)}\right] \approx q^{N}.
  \label{eq:horizon-ideal}
\end{equation}
When inter-cycle transitions are left unregularized, however, the entry distribution drifts and the per-cycle success degrades geometrically, $q_k\approx q\,\gamma^{\,k-1}$ with $\gamma<1$, giving
\begin{equation}
  \Pr\!\left[\text{Succ}_{\text{All}}^{(N)}\right] \approx \prod_{k=1}^{N} q\,\gamma^{\,k-1} = q^{N}\,\gamma^{\,N(N-1)/2},
  \label{eq:horizon-drift}
\end{equation}
an extra $\gamma^{\Theta(N^2)}$ penalty that mirrors the faster-than-geometric collapse of the baselines beyond three objects. Release-and-retreat keeps $\gamma$ closer to one than the baselines, by re-entering the recoverable region after each placement, which moves our policy toward the milder $q^{N}$ regime. This robustness carries over to longer horizons with the same two controllers, with no extra teachers per object, driven by stable inter-stage transitions and broad state coverage rather than task-specific tuning.

\subsection{Robustness and Additional Analysis}
\label{subsec:robustness}

\paragraph{Robustness of the handoff and source of the gain.}
The release-and-retreat handoff depends on only a few switching parameters: a speed/settling threshold, a success threshold, and a timing threshold. Sweeping them over a wide range ($\pm0.5\times$ to $\pm3\times$ the defaults) changes Success~All by at most about $\pm2.3$ points, so the mechanism does not rely on careful manual tuning (full sweep in the supplementary material). We also verify that the benefit comes from the dual-controller structure rather than from extra state coverage alone. A fully random initial state is not a usable alternative: sampled states include penetrations, falls, and invalid contacts. A single controller trained over a broadened initial-state distribution that already includes the post-placement (non-canonical) configurations reaches only $63.7\%$ Success~All (vs.\ $72.4\%$), because the transport and recovery rewards interfere under a single value function.

\paragraph{Role of the language modality.}
Although object/goal state is available during teacher training, the VLA model runs from egocentric RGB and language only. Removing the language token drops Success~All from $63.7\%$ to near zero: under first-person occlusion the policy cannot tell apart visually similar movable objects that share a support surface (\eg, a pot vs.\ a coffee-maker), so language is necessary rather than redundant. The unseen split already includes new language commands, so this also tests instruction generalization.

\subsection{Qualitative Visualization}
\label{subsec:qualitative}

\begin{figure*}[t]
  \centering
  \includegraphics[width=0.78\textwidth]{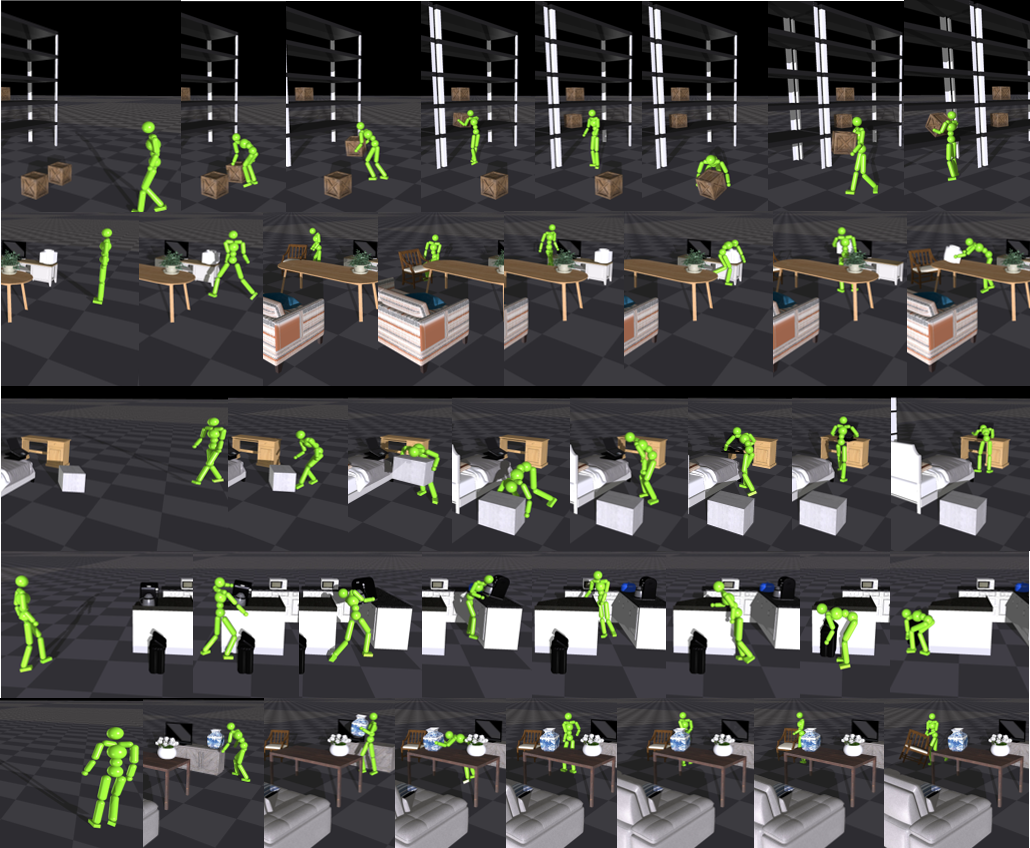}
  \caption{\textbf{Qualitative results across diverse unseen scenes.}
  Each row shows a continuous rollout in a different environment (top to bottom: warehouse, living room 1, bedroom, kitchen, living room 2). The policy produces full long-horizon episodes without resets, demonstrating stable whole-body motion, obstacle-aware navigation, and seamless inter-object transitions.}
  \label{fig:qualitative_results}
\end{figure*}

Figure~\ref{fig:qualitative_results} shows rollout snapshots across five diverse environments, where our policy generalizes to novel furniture and obstacle placements without scene-specific fine-tuning. The character performs stable deep-squat lifts and keeps a wide support polygon under heavier loads (warehouse), navigates narrow passages while re-planning body orientation and arm posture to clear obstacles (bedroom and living rooms), and routes objects around counters with clearance instead of straight-line shortcuts (kitchen).

\subsection{Failure Analysis}
\label{subsec:failure}

Despite strong overall performance, two failure modes persist. On the 350 training tasks, LHM-T fails the first object in $11.2\%$ of episodes and the second after a successful first placement in $16.4\%$; within these slices, grasp/navigation shares are $72\%$/$13\%$ and $54\%$/$36\%$, and the rest is neither. These failures concentrate on flat objects and on narrow, S-shaped passages that are underrepresented in the 350 scenes. Inserting more of these layouts and mining collisions and slipped grasps should cut this tail; the handoff still requires release-and-retreat.

\section{Conclusion}

We presented LHM-Humanoid, a framework and benchmark for producing continuous, reset-free long-horizon whole-body motion, in which a physically simulated humanoid transports multiple objects across cluttered scenes within a single uninterrupted episode. Our key finding is that reset-free composition is governed by recoverability in a two-sided sense: each cycle must end without disturbing the object it just placed, and where the next cycle can begin. Building on this, we integrate learned, viability-aware termination shaping (realized as release-and-retreat), a second controller trained on the induced terminal distribution, and DAgger distillation into one unified policy, which together turn a sequence of contact-rich cycles into a single stable, reset-free rollout. The resulting policy generalizes to unseen scenes and longer sequences and distills into a VLA model, pointing toward a future single recovery-on-demand controller that scales to arbitrarily long horizons without adding controllers. 

\paragraph{Limitations}
The benchmark targets 2--5 objects in physics simulation (Isaac Gym), so longer horizons or transfer to other simulators/morphologies may need adaptation; object and room categories are fixed (no deformables or dynamic obstacles); and the VLA model uses RGB and language only, leaving richer conditioning to future work.

\vspace{\baselineskip}
\noindent\textbf{Acknowledgements.}
This work was partially supported by the ERC Advanced project ETALK (UKRI Funded), the Horizon/UKRI project PRIMI and the US AFOSR project CASPER++.

%
%
\bibliographystyle{splncs04}
\bibliography{main}

@String(JMLR  = {J. Mach. Learn. Res.})

@String(TOG   = {ACM Trans. Graph.})

@String(JMLR  = {JMLR})

@String(TOG   = {ACM TOG})

@inproceedings{hassan2021stochastic,
  title={Stochastic scene-aware motion prediction},
  author={Hassan, Mohamed and Ceylan, Duygu and Villegas, Ruben and Saito, Jun and Yang, Jimei and Zhou, Yi and Black, Michael J},
  booktitle={Proceedings of the IEEE/CVF International Conference on Computer Vision},
  pages={11374--11384},
  year={2021}
}

@article{peng2021amp,
  title={Amp: Adversarial motion priors for stylized physics-based character control},
  author={Peng, Xue Bin and Ma, Ze and Abbeel, Pieter and Levine, Sergey and Kanazawa, Angjoo},
  journal={ACM Transactions on Graphics (ToG)},
  volume={40},
  number={4},
  pages={1--20},
  year={2021},
  publisher={ACM New York, NY, USA}
}

@inproceedings{tessler2023calm,
  title={Calm: Conditional adversarial latent models for directable virtual characters},
  author={Tessler, Chen and Kasten, Yoni and Guo, Yunrong and Mannor, Shie and Chechik, Gal and Peng, Xue Bin},
  booktitle={ACM SIGGRAPH 2023 Conference Proceedings},
  pages={1--9},
  year={2023}
}

@inproceedings{luo2023perpetual,
  title={Perpetual humanoid control for real-time simulated avatars},
  author={Luo, Zhengyi and Cao, Jinkun and Kitani, Kris and Xu, Weipeng and others},
  booktitle={Proceedings of the IEEE/CVF International Conference on Computer Vision},
  pages={10895--10904},
  year={2023}
}

@article{zhu2023neural,
  title={Neural categorical priors for physics-based character control},
  author={Zhu, Qingxu and Zhang, He and Lan, Mengting and Han, Lei},
  journal={ACM Transactions on Graphics (TOG)},
  volume={42},
  number={6},
  pages={1--16},
  year={2023},
  publisher={ACM New York, NY, USA}
}

@article{starke2019neural,
  title={Neural state machine for character-scene interactions},
  author={Starke, Sebastian and Zhang, He and Komura, Taku and Saito, Jun},
  journal={ACM Transactions on Graphics},
  volume={38},
  number={6},
  pages={178},
  year={2019},
  publisher={ACM}
}

@inproceedings{hassan2023synthesizing,
  title={Synthesizing physical character-scene interactions},
  author={Hassan, Mohamed and Guo, Yunrong and Wang, Tingwu and Black, Michael and Fidler, Sanja and Peng, Xue Bin},
  booktitle={ACM SIGGRAPH 2023 Conference Proceedings},
  pages={1--9},
  year={2023}
}

@article{xiao2023unified,
  title={Unified human-scene interaction via prompted chain-of-contacts},
  author={Xiao, Zeqi and Wang, Tai and Wang, Jingbo and Cao, Jinkun and Zhang, Wenwei and Dai, Bo and Lin, Dahua and Pang, Jiangmiao},
  journal={arXiv preprint arXiv:2309.07918},
  year={2023}
}

@inproceedings{pan2024synthesizing,
  title={Synthesizing physically plausible human motions in 3d scenes},
  author={Pan, Liang and Wang, Jingbo and Huang, Buzhen and Zhang, Junyu and Wang, Haofan and Tang, Xu and Wang, Yangang},
  booktitle={2024 International Conference on 3D Vision (3DV)},
  pages={1498--1507},
  year={2024},
  organization={IEEE}
}

@article{wang2023physhoi,
  title={Physhoi: Physics-based imitation of dynamic human-object interaction},
  author={Wang, Yinhuai and Lin, Jing and Zeng, Ailing and Luo, Zhengyi and Zhang, Jian and Zhang, Lei},
  journal={arXiv preprint arXiv:2312.04393},
  year={2023}
}

@article{xie2023hierarchical,
  title={Hierarchical planning and control for box loco-manipulation},
  author={Xie, Zhaoming and Tseng, Jonathan and Starke, Sebastian and van de Panne, Michiel and Liu, C Karen},
  journal={Proceedings of the ACM on Computer Graphics and Interactive Techniques},
  volume={6},
  number={3},
  pages={1--18},
  year={2023},
  publisher={ACM New York, NY, USA}
}

@article{xu2024humanvla,
  title={Humanvla: Towards vision-language directed object rearrangement by physical humanoid},
  author={Xu, Xinyu and Zhang, Yizheng and Li, Yong-Lu and Han, Lei and Lu, Cewu},
  journal={Advances in Neural Information Processing Systems},
  volume={37},
  pages={18633--18659},
  year={2024}
}

@article{wang2024skillmimic,
  title={Skillmimic: Learning reusable basketball skills from demonstrations},
  author={Wang, Yinhuai and Zhao, Qihan and Yu, Runyi and Zeng, Ailing and Lin, Jing and Luo, Zhengyi and Tsui, Hok Wai and Yu, Jiwen and Li, Xiu and Chen, Qifeng and others},
  journal={arXiv e-prints},
  pages={arXiv--2408},
  year={2024}
}

@inproceedings{xu2025intermimic,
  title={Intermimic: Towards universal whole-body control for physics-based human-object interactions},
  author={Xu, Sirui and Ling, Hung Yu and Wang, Yu-Xiong and Gui, Liang-Yan},
  booktitle={Proceedings of the Computer Vision and Pattern Recognition Conference},
  pages={12266--12277},
  year={2025}
}

@inproceedings{pan2025tokenhsi,
  title={Tokenhsi: Unified synthesis of physical human-scene interactions through task tokenization},
  author={Pan, Liang and Yang, Zeshi and Dou, Zhiyang and Wang, Wenjia and Huang, Buzhen and Dai, Bo and Komura, Taku and Wang, Jingbo},
  booktitle={Proceedings of the Computer Vision and Pattern Recognition Conference},
  pages={5379--5391},
  year={2025}
}

@article{ding2025humanoid,
  title={Humanoid-vla: Towards universal humanoid control with visual integration},
  author={Ding, Pengxiang and Ma, Jianfei and Tong, Xinyang and Zou, Binghong and Luo, Xinxin and Fan, Yiguo and Wang, Ting and Lu, Hongchao and Mo, Panzhong and Liu, Jinxin and others},
  journal={arXiv preprint arXiv:2502.14795},
  year={2025}
}

@inproceedings{ross2011reduction,
  title={A reduction of imitation learning and structured prediction to no-regret online learning},
  author={Ross, St{\'e}phane and Gordon, Geoffrey and Bagnell, Drew},
  booktitle={Proceedings of the fourteenth international conference on artificial intelligence and statistics},
  pages={627--635},
  year={2011},
  organization={JMLR Workshop and Conference Proceedings}
}

@inproceedings{huang2023diffusion,
  title={Diffusion-based generation, optimization, and planning in 3d scenes},
  author={Huang, Siyuan and Wang, Zan and Li, Puhao and Jia, Baoxiong and Liu, Tengyu and Zhu, Yixin and Liang, Wei and Zhu, Song-Chun},
  booktitle={Proceedings of the IEEE/CVF Conference on Computer Vision and Pattern Recognition},
  pages={16750--16761},
  year={2023}
}

@inproceedings{jiang2024scaling,
  title={Scaling up dynamic human-scene interaction modeling},
  author={Jiang, Nan and Zhang, Zhiyuan and Li, Hongjie and Ma, Xiaoxuan and Wang, Zan and Chen, Yixin and Liu, Tengyu and Zhu, Yixin and Huang, Siyuan},
  booktitle={Proceedings of the IEEE/CVF Conference on Computer Vision and Pattern Recognition},
  pages={1737--1747},
  year={2024}
}

@inproceedings{zhao2023synthesizing,
  title={Synthesizing diverse human motions in 3d indoor scenes},
  author={Zhao, Kaifeng and Zhang, Yan and Wang, Shaofei and Beeler, Thabo and Tang, Siyu},
  booktitle={Proceedings of the IEEE/CVF international conference on computer vision},
  pages={14738--14749},
  year={2023}
}

@article{jiang2023motiongpt,
  title={Motiongpt: Human motion as a foreign language},
  author={Jiang, Biao and Chen, Xin and Liu, Wen and Yu, Jingyi and Yu, Gang and Chen, Tao},
  journal={Advances in Neural Information Processing Systems},
  volume={36},
  pages={20067--20079},
  year={2023}
}

@article{lin2023motion,
  title={Motion-x: A large-scale 3d expressive whole-body human motion dataset},
  author={Lin, Jing and Zeng, Ailing and Lu, Shunlin and Cai, Yuanhao and Zhang, Ruimao and Wang, Haoqian and Zhang, Lei},
  journal={Advances in Neural Information Processing Systems},
  volume={36},
  pages={25268--25280},
  year={2023}
}

@article{sferrazza2024humanoidbench,
  title={Humanoidbench: Simulated humanoid benchmark for whole-body locomotion and manipulation},
  author={Sferrazza, Carmelo and Huang, Dun-Ming and Lin, Xingyu and Lee, Youngwoon and Abbeel, Pieter},
  journal={arXiv preprint arXiv:2403.10506},
  year={2024}
}

@inproceedings{braun2024physically,
  title={Physically plausible full-body hand-object interaction synthesis},
  author={Braun, Jona and Christen, Sammy and Kocabas, Muhammed and Aksan, Emre and Hilliges, Otmar},
  booktitle={2024 International Conference on 3D Vision (3DV)},
  pages={464--473},
  year={2024},
  organization={IEEE}
}

@inproceedings{cai2021unified,
  title={A unified 3d human motion synthesis model via conditional variational auto-encoder},
  author={Cai, Yujun and Wang, Yiwei and Zhu, Yiheng and Cham, Tat-Jen and Cai, Jianfei and Yuan, Junsong and Liu, Jun and Zheng, Chuanxia and Yan, Sijie and Ding, Henghui and others},
  booktitle={Proceedings of the IEEE/CVF International Conference on Computer Vision},
  pages={11645--11655},
  year={2021}
}

@article{peng2018deepmimic,
  title={Deepmimic: Example-guided deep reinforcement learning of physics-based character skills},
  author={Peng, Xue Bin and Abbeel, Pieter and Levine, Sergey and Van de Panne, Michiel},
  journal={ACM Transactions On Graphics (TOG)},
  volume={37},
  number={4},
  pages={1--14},
  year={2018},
  publisher={ACM New York, NY, USA}
}

@article{peng2022ase,
  title={Ase: Large-scale reusable adversarial skill embeddings for physically simulated characters},
  author={Peng, Xue Bin and Guo, Yunrong and Halper, Lina and Levine, Sergey and Fidler, Sanja},
  journal={ACM Transactions On Graphics (TOG)},
  volume={41},
  number={4},
  pages={1--17},
  year={2022},
  publisher={ACM New York, NY, USA}
}

@inproceedings{juravsky2022padl,
  title={Padl: Language-directed physics-based character control},
  author={Juravsky, Jordan and Guo, Yunrong and Fidler, Sanja and Peng, Xue Bin},
  booktitle={SIGGRAPH Asia 2022 Conference Papers},
  pages={1--9},
  year={2022}
}

@inproceedings{elkoura2003handrix,
  title={Handrix: animating the human hand},
  author={ElKoura, George and Singh, Karan},
  booktitle={Proceedings of the 2003 ACM SIGGRAPH/Eurographics symposium on Computer animation},
  pages={110--119},
  year={2003}
}

@inproceedings{kim2000neural,
  title={Neural network-based violinist's hand animation},
  author={Kim, Junhwan and Cordier, Frederic and Magnenat-Thalmann, Nadia},
  booktitle={Proceedings Computer Graphics International 2000},
  pages={37--41},
  year={2000},
  organization={IEEE}
}

@inproceedings{jain2011controlling,
  title={Controlling physics-based characters using soft contacts},
  author={Jain, Sumit and Liu, C Karen},
  booktitle={Proceedings of the 2011 SIGGRAPH Asia Conference},
  pages={1--10},
  year={2011}
}

@article{zhao2013robust,
  title={Robust realtime physics-based motion control for human grasping},
  author={Zhao, Wenping and Zhang, Jianjie and Min, Jianyuan and Chai, Jinxiang},
  journal={ACM Transactions on Graphics (TOG)},
  volume={32},
  number={6},
  pages={1--12},
  year={2013},
  publisher={ACM New York, NY, USA}
}

@inproceedings{cui2024anyskill,
  title={Anyskill: Learning open-vocabulary physical skill for interactive agents},
  author={Cui, Jieming and Liu, Tengyu and Liu, Nian and Yang, Yaodong and Zhu, Yixin and Huang, Siyuan},
  booktitle={Proceedings of the IEEE/CVF conference on computer vision and pattern recognition},
  pages={852--862},
  year={2024}
}

@inproceedings{dou2023c,
  title={C{\textperiodcentered} ase: Learning conditional adversarial skill embeddings for physics-based characters},
  author={Dou, Zhiyang and Chen, Xuelin and Fan, Qingnan and Komura, Taku and Wang, Wenping},
  booktitle={SIGGRAPH Asia 2023 Conference Papers},
  pages={1--11},
  year={2023}
}

@inproceedings{serifi2024vmp,
  title={Vmp: Versatile motion priors for robustly tracking motion on physical characters},
  author={Serifi, Agon and Grandia, Ruben and Knoop, Espen and Gross, Markus and B{\"a}cher, Moritz},
  booktitle={Computer graphics forum},
  volume={43},
  number={8},
  pages={e15175},
  year={2024},
  organization={Wiley Online Library}
}

@article{yao2024moconvq,
  title={Moconvq: Unified physics-based motion control via scalable discrete representations},
  author={Yao, Heyuan and Song, Zhenhua and Zhou, Yuyang and Ao, Tenglong and Chen, Baoquan and Liu, Libin},
  journal={ACM Transactions on Graphics (TOG)},
  volume={43},
  number={4},
  pages={1--21},
  year={2024},
  publisher={ACM New York, NY, USA}
}

@article{liu2017learning,
  title={Learning to schedule control fragments for physics-based characters using deep q-learning},
  author={Liu, Libin and Hodgins, Jessica},
  journal={ACM Transactions on Graphics (TOG)},
  volume={36},
  number={3},
  pages={1--14},
  year={2017},
  publisher={ACM New York, NY, USA}
}

@article{yang2022learning,
  title={Learning to use chopsticks in diverse gripping styles},
  author={Yang, Zeshi and Yin, Kangkang and Liu, Libin},
  journal={ACM Transactions on Graphics (TOG)},
  volume={41},
  number={4},
  pages={1--17},
  year={2022},
  publisher={ACM New York, NY, USA}
}

@article{yenamandra2023homerobot,
  title={Homerobot: Open-vocabulary mobile manipulation},
  author={Yenamandra, Sriram and Ramachandran, Arun and Yadav, Karmesh and Wang, Austin and Khanna, Mukul and Gervet, Theophile and Yang, Tsung-Yen and Jain, Vidhi and Clegg, Alexander William and Turner, John and others},
  journal={arXiv preprint arXiv:2306.11565},
  year={2023}
}

@article{deitke2022️,
  title={ProcTHOR: Large-Scale Embodied AI Using Procedural Generation},
  author={Deitke, Matt and VanderBilt, Eli and Herrasti, Alvaro and Weihs, Luca and Ehsani, Kiana and Salvador, Jordi and Han, Winson and Kolve, Eric and Kembhavi, Aniruddha and Mottaghi, Roozbeh},
  journal={Advances in Neural Information Processing Systems},
  volume={35},
  pages={5982--5994},
  year={2022}
}

@inproceedings{weihs2021visual,
  title={Visual room rearrangement},
  author={Weihs, Luca and Deitke, Matt and Kembhavi, Aniruddha and Mottaghi, Roozbeh},
  booktitle={Proceedings of the IEEE/CVF conference on computer vision and pattern recognition},
  pages={5922--5931},
  year={2021}
}

@article{wu2023tidybot,
  title={Tidybot: Personalized robot assistance with large language models},
  author={Wu, Jimmy and Antonova, Rika and Kan, Adam and Lepert, Marion and Zeng, Andy and Song, Shuran and Bohg, Jeannette and Rusinkiewicz, Szymon and Funkhouser, Thomas},
  journal={Autonomous Robots},
  volume={47},
  number={8},
  pages={1087--1102},
  year={2023},
  publisher={Springer}
}

@inproceedings{chen2023executing,
  title={Executing your commands via motion diffusion in latent space},
  author={Chen, Xin and Jiang, Biao and Liu, Wen and Huang, Zilong and Fu, Bin and Chen, Tao and Yu, Gang},
  booktitle={Proceedings of the IEEE/CVF conference on computer vision and pattern recognition},
  pages={18000--18010},
  year={2023}
}

@inproceedings{zhao2022compositional,
  title={Compositional human-scene interaction synthesis with semantic control},
  author={Zhao, Kaifeng and Wang, Shaofei and Zhang, Yan and Beeler, Thabo and Tang, Siyu},
  booktitle={European Conference on Computer Vision},
  pages={311--327},
  year={2022},
  organization={Springer}
}

@inproceedings{zitkovich2023rt,
  title={Rt-2: Vision-language-action models transfer web knowledge to robotic control},
  author={Zitkovich, Brianna and Yu, Tianhe and Xu, Sichun and Xu, Peng and Xiao, Ted and Xia, Fei and Wu, Jialin and Wohlhart, Paul and Welker, Stefan and Wahid, Ayzaan and others},
  booktitle={Conference on Robot Learning},
  pages={2165--2183},
  year={2023},
  organization={PMLR}
}

@article{wang2025unified,
  title={Unified Vision-Language-Action Model},
  author={Wang, Yuqi and Li, Xinghang and Wang, Wenxuan and Zhang, Junbo and Li, Yingyan and Chen, Yuntao and Wang, Xinlong and Zhang, Zhaoxiang},
  journal={arXiv preprint arXiv:2506.19850},
  year={2025}
}

@article{driess2025knowledge,
  title={Knowledge insulating vision-language-action models: Train fast, run fast, generalize better},
  author={Driess, Danny and Springenberg, Jost Tobias and Ichter, Brian and Yu, Lili and Li-Bell, Adrian and Pertsch, Karl and Ren, Allen Z and Walke, Homer and Vuong, Quan and Shi, Lucy Xiaoyang and others},
  journal={arXiv preprint arXiv:2505.23705},
  year={2025}
}

@article{li2023object,
  title={Object motion guided human motion synthesis},
  author={Li, Jiaman and Wu, Jiajun and Liu, C Karen},
  journal={ACM Transactions on Graphics (TOG)},
  volume={42},
  number={6},
  pages={1--11},
  year={2023},
  publisher={ACM New York, NY, USA}
}

@article{sutton1999between,
  title={Between MDPs and semi-MDPs: A framework for temporal abstraction in reinforcement learning},
  author={Sutton, Richard S and Precup, Doina and Singh, Satinder},
  journal={Artificial intelligence},
  volume={112},
  number={1-2},
  pages={181--211},
  year={1999},
  publisher={Elsevier}
}

@book{aubin1991viability,
  title={Viability theory: new directions},
  author={Aubin, Jean-Pierre and Bayen, Alexandre M and Saint-Pierre, Patrick},
  year={2011},
  publisher={Springer Science \& Business Media}
}

@inproceedings{pratt2006capture,
  title={Capture point: A step toward humanoid push recovery},
  author={Pratt, Jerry and Carff, John and Drakunov, Sergey and Goswami, Ambarish},
  booktitle={2006 6th IEEE-RAS international conference on humanoid robots},
  pages={200--207},
  year={2006},
  organization={Ieee}
}
\end{document}